\documentclass[sigconf]{acmart}

\usepackage[utf8]{inputenc} 
\usepackage[T1]{fontenc}    
\usepackage{hyperref}       
\usepackage{url}            
\usepackage{booktabs}       
\usepackage{amsfonts}       
\usepackage{amsmath}
\usepackage{nicefrac}       
\usepackage{microtype}      
\usepackage{graphicx}
\usepackage{caption}
\usepackage{mathtools}
\usepackage{subcaption}
\usepackage{wrapfig}
\usepackage{dsfont}
\usepackage{algorithm}
\usepackage{algorithmic}
\usepackage{soul}
\usepackage{enumitem}

\usepackage{xfrac}

\AtBeginDocument{%
  \providecommand\BibTeX{{%
    \normalfont B\kern-0.5em{\scshape i\kern-0.25em b}\kern-0.8em\TeX}}}

\copyrightyear{2020}
\acmYear{2020}
\setcopyright{acmlicensed}
\acmConference[HRI '20]{Proceedings of the 2020 ACM/IEEE International Conference on Human-Robot Interaction}{March 23--26, 2020}{Cambridge, United Kingdom}
\acmBooktitle{Proceedings of the 2020 ACM/IEEE International Conference on Human-Robot Interaction (HRI '20), March 23--26, 2020, Cambridge, United Kingdom}
\acmPrice{15.00}
\acmDOI{10.1145/3319502.3374791}
\acmISBN{978-1-4503-6746-2/20/03}

\begin{document}
\fancyhead{}

\title{Joint Goal and Strategy Inference across Heterogeneous Demonstrators via Reward Network Distillation}

\author{Letian Chen}
\email{letian.chen@gatech.edu}
\affiliation{%
  \institution{Georgia Institute of Technology}
  \city{Atlanta}
  \state{GA}
}

\author{Rohan Paleja}
\email{rpaleja3@gatech.edu}
\affiliation{%
  \institution{Georgia Institute of Technology}
  \city{Atlanta}
  \state{GA}
}

\author{Muyleng Ghuy}
\email{mghuy3@gatech.edu}
\affiliation{%
  \institution{Georgia Institute of Technology}
  \city{Atlanta}
  \state{GA}
}

\author{Matthew Gombolay}
\email{matthew.gombolay@cc.gatech.edu}
\affiliation{%
  \institution{Georgia Institute of Technology}
  \city{Atlanta}
  \state{GA}
}


\begin{abstract}
  Reinforcement learning (RL) has achieved tremendous success as a general framework for learning how to make decisions. However, this success relies on the interactive hand-tuning of a reward function by RL experts. On the other hand, inverse reinforcement learning (IRL) seeks to learn a reward function from readily-obtained human demonstrations. Yet, IRL suffers from two major limitations: 1) \emph{reward ambiguity - } there are an infinite number of possible reward functions that could explain an expert's demonstration and 2) \emph{heterogeneity -} human experts adopt varying strategies and preferences, which makes learning from multiple demonstrators difficult due to the common assumption that demonstrators seeks to maximize the same reward. In this work, we propose a method to jointly infer a task goal and humans' strategic preferences via network distillation. This approach enables us to distill a robust task reward (addressing reward ambiguity) and to model each strategy's objective (handling heterogeneity). We demonstrate our algorithm can better recover task reward and strategy rewards and imitate the strategies in two simulated tasks and a real-world table tennis task. 
\end{abstract}




\maketitle
\vspace{-12pt}
\section{Introduction}
Despite the success reinforcement learning (RL) has achieved in recent years \cite{schulman2017proximal, haarnoja2018soft, finn2016guided, levine2016end}, this optimization technique still requires human RL experts to carefully design reward functions to be effective. Researchers, with the best intentions, can na\"ively specify a reward function that results in an RL policy that catastrophically fails to meet the researcher's latent expectations
\cite{amodei2016concrete}. Unfortunately, the field of RL lacks a standard, repeatable process by which we can specify robotic tasks and have the robot autonomously synthesize sensible policies via RL. Prior research has further shown the difficulty human experts have in defining a desired policy for the robots even though they can readily demonstrate the task \cite{cheng2006feature, raghavan2006active}. As such, researchers have often turned to Imitation Learning (IL), an IRL techniques within Learning from Demonstration (LfD). 

Although IL is a relatively straightforward approach that seeks to directly learn a mapping from states to actions, the learned policy is highly intertwined with environment dynamics. Slight changes in dynamics or highly stochastic environments could cause IL to easily fail due to a ``covariate" shift \cite{ross2011reduction}. In contrast, the demonstrated reward function is a more transferable and robust definition of the task, representing latent objectives a behavior tries to accomplish. Effective reward learning holds utility even after environment dynamics change \cite{daw2014algorithmic}. Recently, researchers have made strides to disentangle reward functions from the environment's dynamics~\cite{fu2017learning}. 

Despite progress, IRL still suffers from fundamental shortcomings~\cite{klein1997recognition}. The first drawback is known as \emph{reward ambiguity}, which is defined as a problem where there exists an infinite number of possible reward functions that could explain an expert's demonstration. The second major drawback is caused by limitations in reasoning about heterogeneous demonstrations. Experts typically adopt heuristics (i.e., ``mental shortcuts") to solve challenging optimization problems, but these highly-refined strategies can present a heterogeneity in behavior across task demonstrators that prior methods fail to effectively reason about. Previous work in IRL has tried to address  heterogeneity \cite{nikolaidis2017game}, but only focused on simplified examples which do not typify the complexity of the real world. 

\begin{figure*}[t!]
  \centering
  \includegraphics[width=0.75\textwidth]{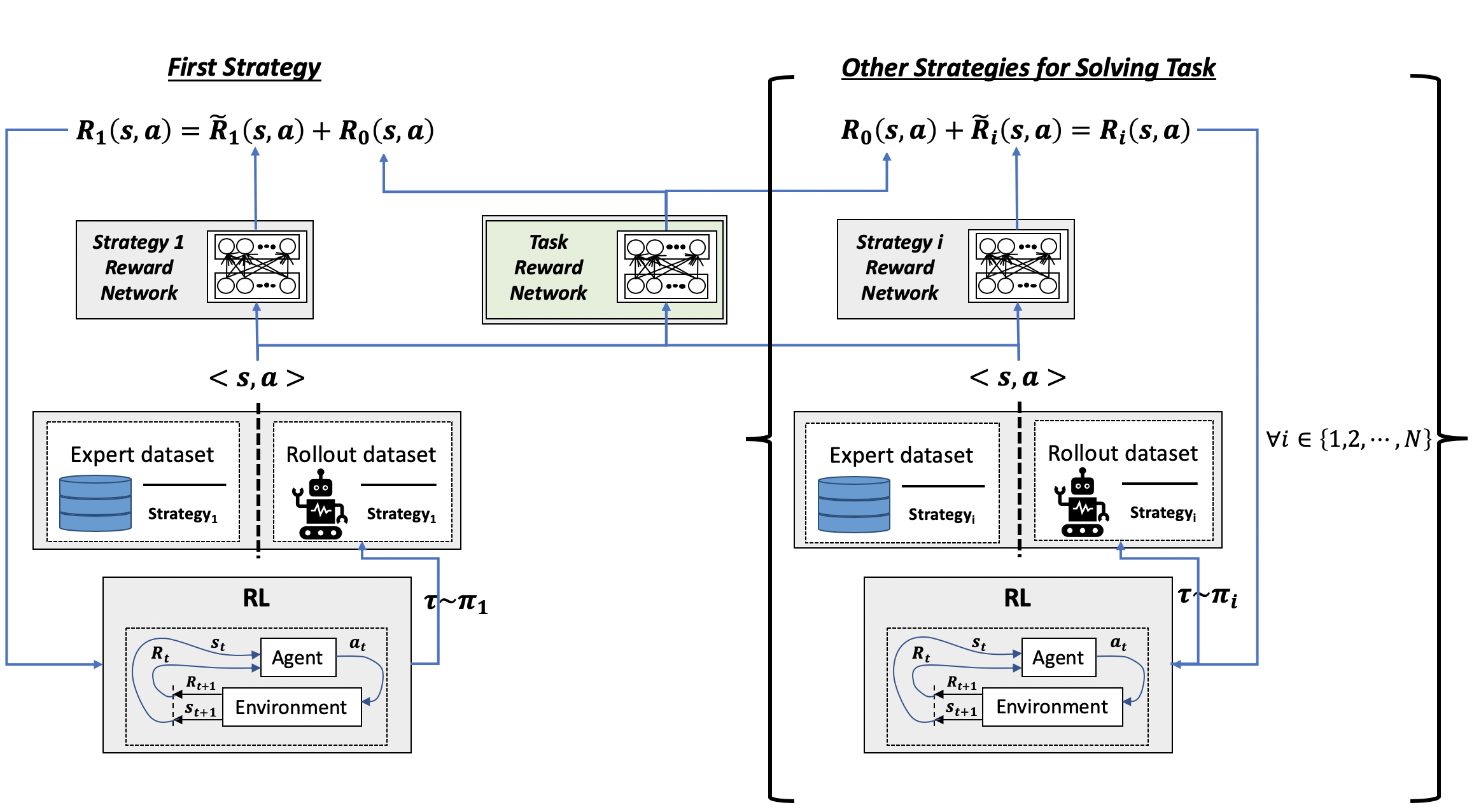}
  \vspace{-12pt}
  \caption{Illustration of Multi-Style Reward Distillation (MSRD). $R_0$ represents task reward, $\widetilde{R_i}$ represents strategy-only reward, and $R_i$ represents strategy-combined reward (see Section \ref{sec:problem_setup} for explanation). Task reward is shared across different strategies, while strategy-only reward is specific to each strategy.}
  \Description{}
  \label{fig:MSRD}
\end{figure*}

To overcome these limitations, we propose Multi-Style Reward Distillation (MSRD, see Figure \ref{fig:MSRD}), a novel IRL technique for learning to tease out task and strategy rewards from heterogeneous demonstration. In MSRD, we model the demonstrated reward function as a combination of the task goal and the demonstrator's strategic preferences. For example, while tennis players seek to win (i.e., the task reward), players exhibit various strategies based on their skills vis-\`{a}-vis those of the opponent (e.g., serve-and-volley, counter-punch, etc). We argue that human experts have specific strategic preferences, especially in complex tasks where it is intractable for humans to compute the global optimum~\cite{klein1997recognition}. In these cases, humans use various heuristics, which results in different types of demonstrations. Unlike prior work in IRL where demonstrators are assumed to optimize a single task reward (i.e., homogeneity), we take on a more human-aware approach by optimizing a combination of the task reward and an intrinsic strategy (i.e., affording heterogeneity).

Our method assumes that expert demonstrators prioritize task completion over conforming to a specific strategy. Under this assumption, heterogeneous demonstrations provide more information to help us tease out the latent task reward versus having only a demonstration of a single strategy. This approach allows us to better model the task reward function by distilling shared components of each demonstrators' trajectories. The task reward recovered from our approach is thus robust and has less ambiguity.

Our contribution in this paper is three-fold:
\begin{enumerate}[topsep=0pt,nolistsep]
    \item We propose a novel IRL framework where we jointly learn task reward and strategic rewards to gain a better estimation of the task reward as well as an estimation for separate strategy reward component for each strategy. 
    \item We show our algorithm's success on two virtual robot control tasks and one real-world physical robot table tennis task. \emph{Nota bene: we are one of few cases that successfully applies IRL on a robotic continuous control problem. }
    \item Our results show that MSRD's learned task reward function achieves high correlation with the ground-truth task reward and that a policy learned from the specific strategy reward recovers the strategic preference. 
\end{enumerate}In addition to MSRD, we also develop a method that helps generate heterogeneous policies of different strategies in simulated environments. This technique is leveraged by our virtual experiments to synthesize heterogeneous task strategies. 

\section{Related Work}

IRL is an LfD approach that aims to infer a demonstrator's objective (i.e., reward function) given a set of performed trajectories. The problem is typically formulated within a Markov Decision Process (MDP) framework. IRL is known to be an ill-posed problem as there are infinitely many reward functions that could explain expert demonstration as optimal including degenerated cases (e.g., $\mathbf{R}=0$ \cite{ng2000algorithms}). Two primary methods exist in the IRL literature that aim to solve this ambiguity: maximum margin approaches \cite{abbeel2004apprenticeship, ratliff2006maximum} and probabilistic approaches \cite{ramachandran2007bayesian, ziebart2008maximum, ziebart2010modeling}. Maximum margin approaches try to find a reward function that explains an expert's trajectory not only as being optimal, but also as being better than all other trajectories by a margin. Probabilistic approaches (e.g. Bayesian IRL \cite{ramachandran2007bayesian}, Maximum-Entropy IRL \cite{ziebart2008maximum}, and Maximum Causal Entropy IRL \cite{ziebart2010modeling}) assume trajectories with higher rewards have an exponentially higher probability of being generated and apply a maximum likelihood framework to solve for such a reward function. Advantages of this framework include its tolerance of non-optimal demonstrations and ease of applying gradient-based optimization. Maximum-entropy based IRL methods are thus suitable for deep neural network models to find the reward function \cite{wulfmeier2015maximum}, obviating the need for feature engineering. Despite the advantages of a maximum entropy IRL (ME-IRL) framework, perfect information about the system dynamics and the ability to calculate the exact state feature counts are required. Guided Cost Learning (GCL \cite{finn2016guided}) and its successor, Adversarial Inverse Reinforcement Learning (AIRL \cite{fu2017learning}) have been proposed to tackle ME-IRL problems. AIRL poses the reward learning problem in an adversarial setting, where the reward function tries to assign a high reward to expert demonstrations while assigning a low reward to generated trajectories. Meanwhile, the policy is optimized over the learned reward function to maximize its expected reward. 

Alternatively, preference learning is another well-explored research area that aims to learn individuals' proclivities. In robotics, preference learning typically involves learning a human's preference and adapting a robot's behavior to better collaborate with human-beings \cite{nikolaidis2017game,gombolay2018robotic}. There are several methods that could infer preference information without asking the human to explicitly provide preference information~\cite{gombolay2016apprenticeship,gombolay2018human}. For example, \citet{schafer2007collaborative} utilizes collaborative filtering for a recommender system and \citet{xu2018learning} formulates a meta-IRL problem that could learn a prior over preferences. A more recent direction of preference learning lies in the multi-task setting. \citet{dimitrakakis2011bayesian} models reward-policy pairs $(R_i, \pi_i)$ drawn from an unknown prior. \citet{choi2012nonparametric} integrates Dirichlet process mixture model into Bayesian IRL as a task prior. Repeated inverse reinforcement learning (RIRL) \cite{amin2017repeated} formalizes the setting in which a user is observed performing different tasks. The goal of RIRL is to infer a task-independent preference. However, one of the biggest limitations of RIRL is the method assumes full knowledge of each task's reward function and the relative weights between task rewards and preference rewards, which is unrealistic in most real-world applications. Observational Repeated IRL (ORIRL) relaxes the assumption by introducing a learnable relative weight yet still assumes perfect knowledge of the task reward~\citet{woodworth2018preference} . Furthermore, both RIRL and ORIRL work is carried out in discrete finite state-action space, which is a significant simplification of real-world, continuous, state-action space. While preference learning in a multi-task setting tries to learn shared preferences across tasks, we propose to learn a shared task reward from observing different strategies for the same task. 

Learning from heterogeneous demonstrations poses a significant challenge in LfD. The typical approach is to assume homogeneity over demonstrators; however, [32]  found that commercial line pilots exhibit significant heterogeneity in performing a well-posed task (e.g., executing a pre-specified flight plan) as to make it more practical to learn from a single trajectory and disregard the remaining data. We are motivated by such settings to develop a method that might still be capable of generating a robust task specification while also leveraging all available data to help mitigate the curse of dimensionality.

Few have addressed multiple-reward-function learning. One such approach by \citet{nikolaidis2014efficient} clusters different kinds of behaviors and applies inverse reinforcement learning on each cluster. Another approach, Option GAN \cite{henderson2018optiongan}, learns a division in demonstration state space and a separate reward function and policy for each subspace. However, neither considers the relationship between the learned reward functions. In contrast, our work poses a common task reward function in each of the reward functions, together with a separate strategy reward function for each strategy, which enables us to tease out task reward and strategy rewards. 

Another line of work that tries to model heterogeneous expert data lies in imitation learning. Generative Adversarial Imitation Learning (GAIL) \cite{ho2016generative} is a popular imitation learning algorithm in which a discriminator tries to distinguish between expert trajectories and generated trajectories, while a generator tries to deceive the discriminator. Some extensions to GAIL utilize a latent variable to model multi-modal (multi-strategy) demonstration data \cite{hausman2017multi, li2017infogail}. Despite the good performance of imitation learning algorithms in some contexts, such methods are severely sensitive to a change of environments (e.g., a small disturbance could result in total failure of intended task). By contrast, the reward function is more robust to a change of dynamics in the environment, as the function represents the intention of the behavior. In this work, we separate the task reward and the strategy reward, thereby better extracting a transferable signal describing the demonstrators' latent intent. 

In contrast, our approach proposes a novel problem setup that considers multi-strategy demonstrations for a single task. We propose a distillation-regularization structure between the task reward and strategy rewards, shown in Figure \ref{fig:MSRD}, which aims to distill common task rewards and separate strategy preferences. \textcolor{black}{Our demonstration of MSRD for robot table tennis was inspired by \citet{peters2013towards}, who proposed  robot table tennis as a challenging task for robot learning. In a series of papers, these researchers designed sophisticated models, features~\cite{muelling2014learning}, and rewards~\cite{mulling2013learning} to train versatile policies for robot skill learning~\cite{muelling2010learning,KOC2018121}.}

\section{Preliminaries}
\subsection{Markov Decision Process}

An MDP, $M$, is a 6-tuple $(S,A,R,T,\gamma,\rho_0)$, where $S$ and $A$ are the state and action spaces, respectively. $\gamma\in (0,1)$ is the temporal discount factor. $R(s,a)$ represents the reward after executing action $a$ in state $s$. In some cases, $R(s,a)$ could be simplified as $R(s)$. $T(s,a,s^\prime)$ is the transition probability to $s^\prime$ from state $s$ after taking action $a$. $\rho_0(s)$ is the initial state distribution. The standard goal of reinforcement learning is to find the optimal policy $\pi^*$ that maximizes the discounted future reward, $J(\pi)=\mathbb{E}_{\tau\sim \pi}\left[\sum_{i=0}^T{\gamma^tR(s_t,a_t)}\right]$, where $\tau=(s_0,a_0,\cdots,s_T,a_T)$ denotes a sequence of states and actions induced by the policy and dynamics. We instead consider a more general maximum entropy objective introduced by \citet{ziebart2010modeling}, which augments the standard objective with an entropy bonus to favor stochastic policies and to encourage exploration during optimization,
$J(\pi)=\mathbb{E}_{\tau\sim \pi}\left[\sum_{i=0}^T{\gamma^tR(s_t,a_t)}+\alpha H(\pi(\cdot | s_t))\right]$.

\subsection{Inverse Reinforcement Learning}
IRL considers an MDP sans reward function ($M\backslash R$) with the goal being to infer reward function $R(s,a)$ given a set of demonstrated trajectories $\mathcal{D}=\{\tau_1, \tau_2, \cdots, \tau_N\}$. A typical assumption for IRL is that demonstrated trajectories are optimal, or at least near-optimal. In the maximum entropy IRL (Max-Ent IRL) framework,  inference of the reward function is turned to a maximum likelihood optimization problem by assigning occurrence probability of a trajectory proportional to the exponential of discounted cumulative reward, $p_\theta(\tau)\propto e^{\sum_{t=0}^T{\gamma^tr_\theta(s_t,a_t)}}$. Therefore, Max-Ent IRL aims to find the reward function under which the demonstrated trajectories have the highest likelihood, as shown in Equation \ref{eqn:maximum_entropy_irl}.
\begin{align}
\label{eqn:maximum_entropy_irl}
    \max_\theta{\mathbb{E}_{\tau\sim \mathcal{D}}[\log p_\theta(\tau)]}
\end{align}
AIRL \cite{fu2017learning} casts the optimization in a generative adversarial framework, learning a discriminator, $D$, to distinguishes between experts and a deceiving generator, $\pi$, that learns to imitate the expert. This framework follows Max-Ent IRL's assumption that trajectory likelihood is proportional to the exponential of rewards. $D$ is defined in Equation \ref{eqn:D_and_f}, where $f_\theta(\tau)$ is the learnable reward function and $\pi$ is the current policy. $D$ is updated by minimizing its cross entropy loss to distinguish expert trajectories from generator policy rollouts (Equation \ref{eqn:airl_loss}).  $\pi$ is trained to maximize the pseudo-reward function given by $f(s,a)$. 
\begin{equation}
    D_\theta(s,a)=\frac{\exp(f_\theta(s,a))}{\exp(f_\theta(s,a))+\pi(a|s)}
    \label{eqn:D_and_f}
\end{equation}

\begin{align}
    L_D=-\mathbb{E}_{\tau\sim\mathcal{D}, (s,a)\sim\tau}[&\log D(s,a)] \nonumber \\
    &-\mathbb{E}_{\tau\sim\pi, (s,a)\sim\tau}[1-\log D(s,a)]
\label{eqn:airl_loss}
\end{align}

\subsection{Neural Network Distillation}
Neural network distillation applies supervised regression to train a student network to produce the same output distribution as a trained teacher network. The method was first proposed by \citet{hinton2015distilling} and has been applied in RL mainly for performing policy distillation \cite{rusu2015policy, teh2017distral, czarnecki2019distilling}. Particularly, \citet{teh2017distral} proposes that instead of distilling each policy $\pi_i$ to a general policy $\pi_0$, we could gain a faster convergence with a two-column architecture by defining $\pi_i=\pi_0 + \widetilde{\pi_i}$. Accordingly, $\pi_i$ only needs to learn a near-zero difference between a common policy and the task-specific policy. 

\section{Method}
\subsection{Problem Setup}
\label{sec:problem_setup}
We consider a setup in which there is only one task (one MDP $M=(S,A,R^{(0)},T,\gamma,\rho_0)$), but the demonstrations are generated by employing varying strategies. Different strategies may come from different experts who have personalized strategical preferences or one expert that has mastered several. Therefore, despite heterogeneity within demonstrations, all trajectories in the dataset should still be near-optimal in terms of task reward $R^{(0)}$. We denote each strategy's demonstration dataset as $\mathcal{D}^{(i)}=\{\tau_1^{(i)},\tau_2^{(i)}, \cdots, \tau_M^{(i)}\}$, where $i\in\{1,2,\cdots, N\}$ is strategy index, $M$ is the number of demonstration trajectories for one strategy, and $N$ is the number of strategies. Our first objective is to infer a shared task reward function $R^{(0)}$ despite there being different strategies in the demonstration dataset. The second objective is to infer the strategy-only reward $\widetilde{R^{(i)}}$. Combining strategy-only reward with the task reward function will result in a strategy-combined reward $R^{(i)}$, which should induce the observed expert strategical behaviors. 

\subsection{Task and Strategy Reward}
\label{sec:reparameterization}

We first propose to model the strategy-combined reward function that is optimized by a demonstrator to be a linear combination of the task and the strategy-only reward as given by Equation \ref{eqn:reparameterization}.
\begin{align}
    R^{(i)}(\cdot) &= R^{(0)}(\cdot) + \alpha_i \widetilde{R^{(i)}}(\cdot) 
    \label{eqn:reparameterization}
\end{align}

Despite the simplicity, we argue Equation \ref{eqn:reparameterization} makes a reasonable assumption. Several previous IRL works also apply linear, if not more constraining, assumptions to combine reward functions: \citet{amin2017repeated} create a combined reward by adding a task reward with a cross-task shared preference, while we add shared task reward with each strategy reward; \citet{woodworth2018preference} propose a similar formulation, except their task reward is known and they try to learn task-independent preference in a multi-task setting. In the RL literature, there is also substantial work combining rewards linearly (e.g., adding intrinsic reward, such as curiosity or entropy to the original task reward for the sake of exploration). In fact, many engineered reward functions are also a linear combination of several reward components (e.g., for OpenAI Gym \cite{1606.01540} MuJoCo \cite{todorov2012mujoco} hopper environment, its reward function is defined as a linear combination of forward speed, living bonus, and action penalty). Additionally, assuming a linear combination could also provide interpretability (see Figure \ref{fig:ip_reward_f}). Our approach is unique in that it shares the task reward while having the flexibility on each strategy reward, which we argue is more realistic while allowing us to apply joint inference of the task reward and strategy reward. 

\subsection{Reward Network Distillation}

To infer the shared task reward function between different strategies, we propose utilizing network distillation to distill common knowledge from each separately learned strategy-combined reward $R^{(i)}$ to the task reward function $R^{(0)}$. We also want to regularize $R^{(i)}$ to be close to $R^{(0)}$, since we have the prior knowledge that despite heterogeneous strategic preferences, all experts should still be prioritizing optimizing the task reward $R^{(0)}$ to achieve at least near-optimal performance. Previous distillation methods mainly focus on distilling classification results, and therefore KL-divergence between teacher and student outputs could be a good choice for regularization. However, reward functions are real-valued functions and therefore probabilistic distance metrics do not fit. Thus, we propose to regularize the expected L2-norm of the difference between the reward functions, as shown in Equation \ref{eqn:l2}, in which $\pi^{(i)}$ is the optimal policy under reward function $R_{\theta_i}^{(i)}$. 
\begin{align}
\label{eqn:l2}
    L_\text{reg}=\mathbb{E}_{(s,a)\sim \pi^{(i)}}\left({\left\lvert\left\lvert  R_{\theta_i}^{(i)}(s,a)-R_{\theta_0}^{(0)}(s,a)\right\rvert\right\rvert}_2\right)
\end{align}
Note that we are using an index both on $\theta$ and $R$ to denote that each strategy-combined reward $R^{(i)}$ has its own reward parameters, and that these are approximated by separate neural networks with parameters $\theta_i$ for each strategy and $\theta_0$ for the task reward. There is no parameter sharing between strategy and task reward. 

Due to the computational cost of optimizing $\pi^{(i)}$ using RL, we seek to avoid fully optimizing it inside the IRL loop. Therefore, we apply an iterative reward function and policy training schedule, similar to AIRL. Through such way, we could learn reward function and policy simultaneously. Combining AIRL's objective (Equation \ref{eqn:airl_loss}) with our distillation objective, we maximize $L_D$ in Equation \ref{eqn:vanilla_distillation}. 
\begin{align}
\begin{split}
\label{eqn:vanilla_distillation}
    L_D=\sum_{i=1}^N&\left[{\mathbb{E}_{(s,a)\sim \tau_j^{(i)}\sim \mathcal{D}^{(i)}}{\log D_{\theta_i}(s,a)}}  \right.\\
    &\indent\indent+ {\mathbb{E}_{(s,a)\sim \pi^{(i)}}{\log (1-D_{\theta_i}(s,a))}}\\ 
    &\indent\indent\indent\indent- \left.{\mathbb{E}_{(s,a)}\left({\left\lvert\left\lvert R_{\theta_i}^{(i)}(s,a)-R_{\theta_0}^{(0)}(s,a)\right\rvert\right\rvert}_2\right)}\right]
\end{split}
\end{align}
$D$ is dependent on $\theta_i$ via Equation \ref{eqn:D_and_f} ({\small$R_{\theta_i}^{(i)}$} corresponds to $f_\theta$). Each $\pi^{(i)}$ optimizes\footnote{We choose Trust Region Policy Optimization (TRPO)~\cite{fu2017learning}} {\small$R_{\theta_i}^{(i)}$}. Yet, while Equation \ref{eqn:vanilla_distillation} should be able to distill the shared reward into {\small$R_{\theta_0}^{(0)}$}, the distillation is inefficient as {\small$R_{\theta_0}^{(0)}$} will work as a strong regularization for {\small$R_{\theta_i}^{(i)}$} before successful distillation. 

Instead, our structure in Equation \ref{eqn:reparameterization} allows for a two-column re- parameterization, speeding up knowledge transfer and making the learning process easier \cite{teh2017distral}. Combining Equation \ref{eqn:reparameterization} and Equation \ref{eqn:vanilla_distillation}, we arrive at Equation \ref{eqn:reparameterized_distillation}.

\begin{align}
\label{eqn:reparameterized_distillation}
    L_D&=\sum_{i=1}^N\left[{\mathbb{E}_{(s,a)\sim \tau_j^{(i)}\sim \mathcal{D}^{(i)}}{\log D_{\theta_i,\theta_0}(s,a)}}\right. \nonumber\\
    &\indent\indent+ {\mathbb{E}_{(s,a)\sim \pi^{(i)}}{\log \left(1-D_{\theta_i,\theta_0}(s,a)\right)}} \nonumber \\
    &\indent\indent\indent\indent- \left.\alpha_i{\mathbb{E}_{(s,a)}\left({\left\lvert\left\lvert \widetilde{R_{\theta_i}^{(i)}}(s,a)\right\rvert\right\rvert}_2\right)}\right]
\end{align}

The key difference between Equation \ref{eqn:reparameterized_distillation} and Equation \ref{eqn:vanilla_distillation} is that $D$ depends on both {\small$R_{\theta_0}^{(0)}$} and {\small$\widetilde{R_{\theta_i}^{(i)}}$} instead of separate $R_{\theta_i}^{(i)}$. Thus, {\small $R_{\theta_0}^{(0)}$} directly updates from the discriminator's loss rather than waiting for knowledge to be learned by a strategy-combined reward and subsequently distilled into a task reward. Further, the last term of Equation \ref{eqn:reparameterized_distillation} reduces to a simple L2-regularization on strategy-only reward's output, weighted by $\alpha_i$. This formulation provides us with a new view to interpret the relative weights of the strategy-only reward $\alpha_i$: the larger $\alpha_i$ is, the more the strategy-only reward will influence the strategy-combined reward. Therefore, we will have higher regularization to account for possible overwhelming of the task reward function. Comparing Equation \ref{eqn:reparameterized_distillation} and \ref{eqn:airl_loss}, we could interpret MSRD in another view: optimizing $\theta_i$ only via IRL objective results in a combination of task and strategy reward, and adding regularization on strategy reward will encourage to encode only necessary information in $\theta_i$ and share more knowledge in $\theta_0$.

\subsection{Multi-Style Reward Distillation}
We summarize our algorithm in Algorithm \ref{algo:msrd}. In the algorithm we first collect the heterogeneous-strategy expert dataset and initialize network parameters as well as relative weights. For each training epoch, we will iterate over all strategies (line 5). For each strategy, we first collect $K$ trajectories generated by its corresponding policy $\pi^{(i)}$ (line 6). We also sample expert trajectories for strategy  $i$ from the dataset (line 7). We then train the Discriminator (reward function) with loss given by Equation \ref{eqn:reparameterized_distillation} and data $\tau_j^{\text{gen}}$ and $\tau_j^{\text{exp}}$ (line 8).  After training the reward function, we could assign pseudo-reward to trajectories we generate with reward function $R_{\theta_i}^{(i)}$ (line 9). Finally, we update the policy according to the trajectory generated and pseudo-reward signal (line 10). In practice, we could also postpone the gradient update for $R^{(0)}$ at the end of one sweep of strategies to stabilize the learning of the task reward. 
\begin{algorithm}[t!]
\caption{Multi-Strategy Reward Distillation}
\label{algo:msrd}
\begin{algorithmic}[1]
\STATE Obtain heterogeneous-strategy expert dataset $\mathcal{D}^{(i)}=\{\tau_1^{(i)},\cdots, \tau_M^{(i)}\}$ \ $\forall i \in \{1,\cdots,N\}$
\STATE Initialize $R^{(0)}$, $R^{(i)}$, $\pi^{(i)}$\quad$\forall i \in \{1,2,\cdots,N\}$
\STATE Determine relative weights $\alpha_i\quad\forall i \in \{1,2,\cdots,N\}$
\WHILE{not converged}
 \FOR{i = 1 to $N$}
    \STATE Collect $K$ ($K \leq M$) trajectories $\tau_j^{\text{gen}}$ by executing $\pi^{(i)}$
    \STATE Sample $K$ trajectories $\tau_j^{\text{exp}}$ from $\mathcal{D}^{(i)}$
    \STATE Train $\theta_i$ and $\theta_0$ with Equation \ref{eqn:reparameterized_distillation}, $\tau_j^{\text{gen}}$ and $\tau_j^{\text{exp}}$
    \STATE Assign reward for all transitions in $\tau_j^{\text{gen}}$: $r(s,a)=R_{\theta_i}^{(i)}(s,a)$
    \STATE Update policy $\pi_i$ using trajectories $\tau_j^{\text{gen}}$ via TRPO with entropy bonus to encourage exploration
 \ENDFOR
 \ENDWHILE
 \RETURN $R^{(0)}$, $R^{(i)}$, $\pi^{(i)}$
\end{algorithmic}
\end{algorithm}
\begin{figure*}[t]
\centering
\begin{minipage}{.625\textwidth}
  \centering
  \includegraphics[width =\linewidth, height = 5cm]{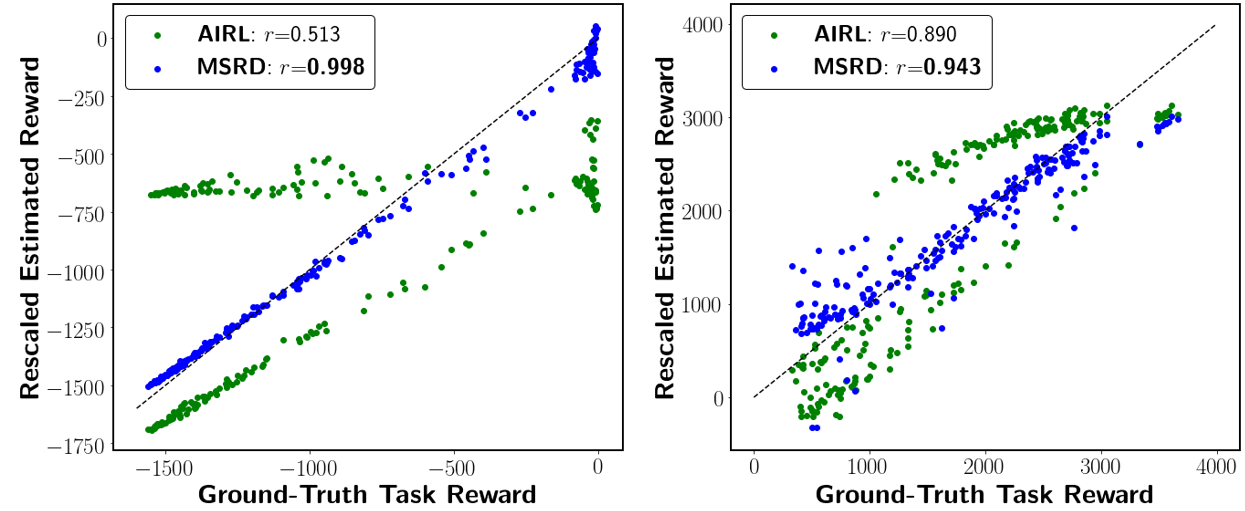}
  \captionof{figure}{Correlation between ground-truth and estimated task reward, normalized for each strategy to $[0,1]$, for inverted pendulum (left) and hopper (right) environments. Reward is invariant to shift/scale. $r$ is correlation coefficient.}
  \label{fig:corr}
\end{minipage}%
\hspace{.5cm}
\begin{minipage}{.325\textwidth}
  \centering
  \includegraphics[width = \linewidth, height = 5cm]{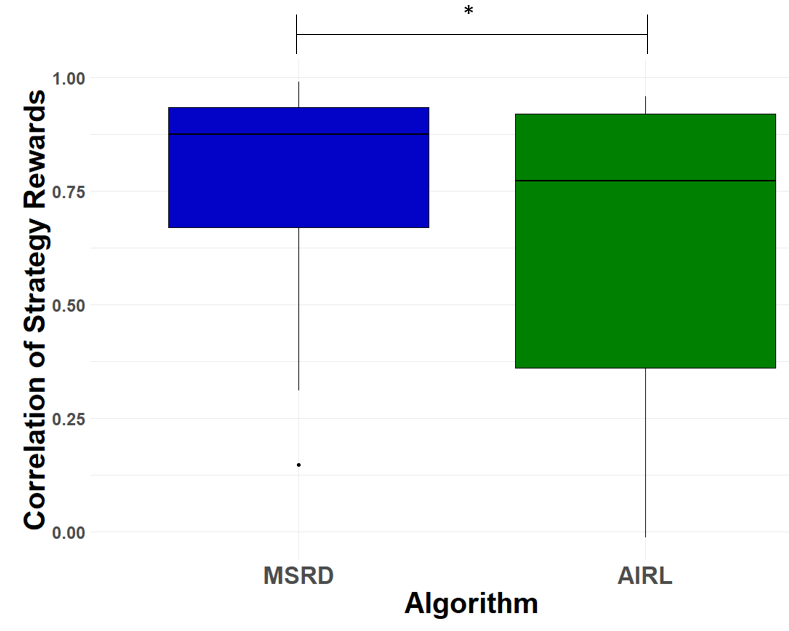}
  \captionof{figure}{Correlation between ground-truth vs.~estimated strategy reward by MSRD and AIRL on Inverted Pendulum.}
  \label{fig:strategy_corr}
\end{minipage}
\end{figure*}

\section{Experiment Setup}
We tested MSRD on both virtual and real-world environments. Here, we describe our experiments, showing results in Section \ref{sec:results}. 

\subsection{Virtual Experiments}

We first tested MSRD on two simulated environments: a simpler inverted pendulum control task and a more difficult hopper locomotion task (see Figure \ref{fig:hopper_strategy_reward} and supplementary materials for illustration). The goal of the inverted pendulum task is to balance a pendulum on a cart by moving the cart left/right, making it a single degree of freedom (DoF) problem, based on a 2D observation. The reward for inverted pendulum is defined as the negative absolute value of the pendulum angle from upright position. The objective of hopper is to control 3-DoF joints to move forward based on its 11-dimensional observation. The reward for hopper is defined as the speed at which it moves forward. We used the OpenAI Gym \cite{1606.01540} MuJoCo \cite{todorov2012mujoco} implementation for both environments but made the following changes to fit our application: 1) remove termination judgements to gain flexibility in behaviors; 2) add timeout constraint of 1,000 steps. 

\subsubsection{Heterogeneous Demonstration Collection}
Our algorithm can utilize heterogeneous-strategy demonstrations. Therefore, we first need to generate a variety of demonstrations to emulate heterogeneous strategies that humans will apply in solving problems for our virtual experiments. Typical RL algorithms can only generate single-modal policy for each task in attempt to maximize the task reward. Some previous work has tried to generate diverse behaviors \cite{eysenbach2018diversity, zhang19q}. Among them, Diversity is All You Need (DIAYN) trains a discriminator to distinguish different behaviors and trains policies utilizing the discriminator's output as a pseudo-reward. The pseudo-reward is shown in Equation \ref{eqn: pseudo_reward}, where $z$ is the strategy index, $q$ is the posterior decoding $z$ from $s$, and $p(z)$ is the prior distribution. However, DIAYN only discovers different behaviors without task specification, so we augment DIAYN to incorporate task reward by the linear form of Equation \ref{eqn:reparameterization}. 
\begin{align}
    r_z(s,a)=\log q_\phi(z|s)-\log p(z), 
    \label{eqn: pseudo_reward}
\end{align}

We also propose a method to encourage different strategies taking different actions in the same state via a diversity reward in which $k$ is the strategy index and $\pi_k$ is the policy for strategy $k$ (Equation \ref{eqn:kle}). Equation \ref{eqn:kle} encourages the KL-divergence between different strategy's policies to be large. Linearly combining KL-encouraged diversity reward with the task reward, we can train strategies that optimize both the task goal and the diversity goal. 
\begin{align}
    r_\text{KL}^k(s)=\sum_{i=1}^N{KL(\pi_k(\cdot |s)|| \pi_i(\cdot | s))},
    \label{eqn:kle}
\end{align}
We trained both ``DIAYN + Extrinsic Reward" and "KL-Encouraged + Extrinsic Reward" policies to collect heterogeneous trajectories that applied different strategies to solve the task. From all the strategies generated, we chose 20 strategies for inverted pendulum and the two most significant strategies for hopper. Generally, different strategies in inverted pendulum encourage the cart to stay at different angles, but some strategies maintain dynamic balance by periodically moving the cart left and right. Two different strategies in hopper are "Hop" and "Crawl" as illustrated in Figure \ref{fig:hopper_strategy_reward}. More details are available in the supplementary materials.

\subsection{Real-World Experiments}
\label{sec:pingpong_env}
For our second environment, we tested MSRD on its capability to learn various table tennis strokes from expert human demonstration. Participants were asked to kinetically teach a robot arm to hit an incoming ping pong ball using three different stroke strategies: push, slice, and top spin (see supplementary videos for illustration), from both a forehand and backhand position. Detailed environment setup could be found in supplementary.

\subsubsection{Experiment Design and Subjects}
We adopt a within-subjects design for demonstration collection, requiring subjects finish all six combinations of positions and strategies. We pseudo-randomized the order of forehand and backhand, as well as the order of three strategies. We recruited 10 subjects from a population of college graduate students. Subjects were compensated.

\subsubsection{Experiment Procedure}
When the participants arrived, they read and signed a consent form detailing the purpose, duration, and study procedure. For each trial, we began by showing a video tutorial on how to move the robot arm to hit the incoming ball with a specific strategy. After the tutorial, we allowed participant to practice with the automatic ping pong feeder. Once the subject felt ready, we began the recording process. Saved trajectories were those in which the participant was able to successfully return the ball to the opponent's side and the movement of striking closely resembled the strategy assigned. We collected three recordings for each conditions, position (forehand/backhand), and strategy combination. We record at each timestep of the trajectory the robot joint angles, angle rates and the ball position. 

\section{Results and Discussion}
\label{sec:results}

In this section, we report and analyze MSRD's results on three environments and benchmark against AIRL to elucidate MSRD's advantage on recovering both the latent task reward (the essential goal of the demonstrators) and the means by which the task is accomplished (i.e. the strategy). We explore two hypotheses:

\noindent\textbf{H1:} \emph{The task reward learned by MSRD has a higher correlation with the true task reward than AIRL.}

\noindent\textbf{H2:} \emph{Strategy-only reward learned by MSRD has a higher correlation with true strategic preferences than AIRL.}

We assessed both hypotheses quantitatively and qualitatively for the simulation environments only as the ground-truth reward functions are available. In the physical robot experiment, users are instructed to execute the task instead of optimizing an objective function, meaning that we do not have access to the underlying task reward or strategy-only rewards. Therefore, \textbf{H1} and \textbf{H2} were assessed qualitatively in the physical robot experiment. We did not compare MSRD method with more traditional IRL methods (such as Max-Margin IRL, Feature-Matching IRL, Max-Ent IRL, etc.) because they rely on sophisticated feature designs. We leverage end-to-end neural network architectures to avoid manual feature extraction. Furthermore, we wanted to provide a fair comparison between methods; given MSRD leverages AIRL, the most appropriate choice is thus AIRL. In future work, we propose considering integrated other IRL methods into MSRD. In future work, our method could be combined with any existing strategy inference method (e.g., \cite{nikolaidis2014efficient}) to make it applicable to cases where strategy labels are unavailable.

\begin{figure}[t]
  \centering
  \includegraphics[width=0.9\linewidth, height = 4cm]{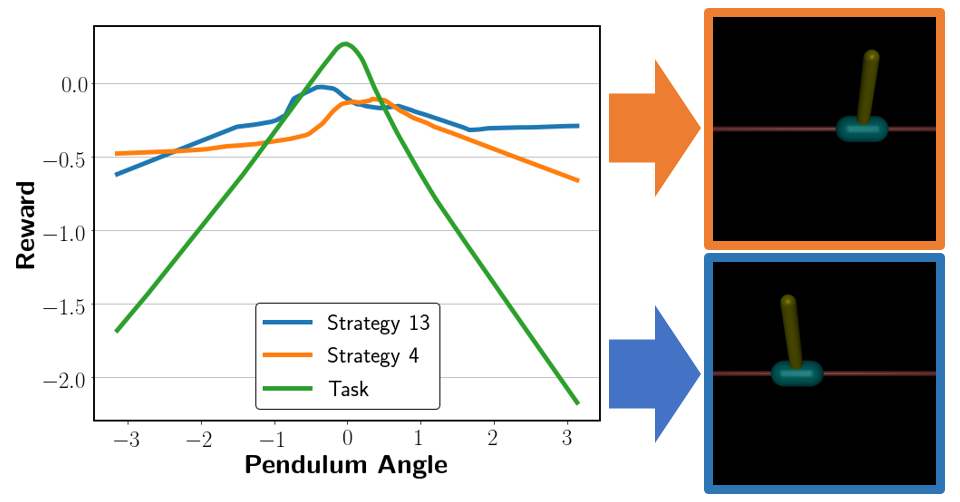}
  \caption{Task/Strategy Reward Functions of Inverted Pendulum vs Pendulum Angle and Corresponding Behaviors.}
  \Description{}
  \label{fig:ip_reward_f}
\end{figure}

\subsection{Simulated Environments}
To test \textbf{H1}, we constructed a dataset of trajectories that have various task performances utilizing noise injection \cite{brown2019ranking}. We note that this dataset was also generated with various-strategy policies to be representative of the entire trajectory space. We then evaluated the reward function learned by AIRL and MSRD on the trajectories, comparing estimated vs.~ground-truth rewards. We show a correlation of estimated rewards and ground-truth task rewards in Fig.\ref{fig:corr}. The task reward function learned through MSRD has a higher correlation with the ground-truth reward function (0.998 and 0.943) versus AIRL (0.51 and 0.89) for each domain, respectively). AIRL's reward function overfits to some strategies and mixes the task reward with that strategy-only reward, making its estimation unreliable for other strategies' trajectories. 

\begin{figure}[t]
  \centering
  \includegraphics[width=0.8\linewidth]{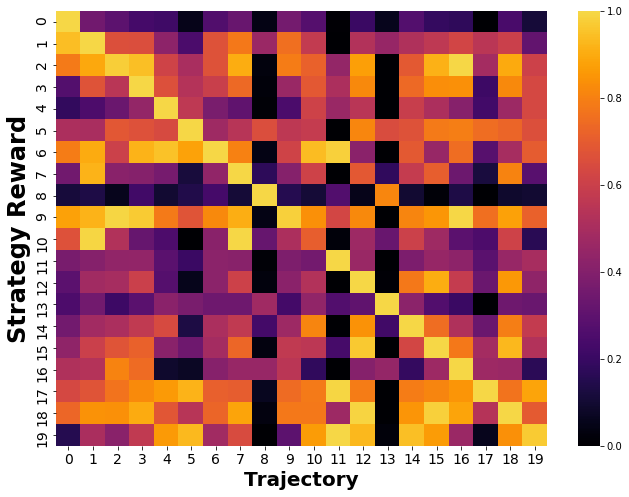}
  \caption{Evaluation of strategy rewards on strategy trajectories (Inverted Pendulum); rows normalized to $[0,1]$. }
  \Description{}
  \label{fig:ip_heatmap}
\end{figure}

To test \textbf{H2}, we calculated the correlations of MSRD's strategy-only rewards with the true strategic preferences and compared that with the correlation of AIRL's rewards when AIRL is trained on each individual strategy. In simulated domains, true strategic preferences are available as the pseudo-reward in Equation \ref{eqn: pseudo_reward} and \ref{eqn:kle}. Correlations of both methods for all strategy rewards in inverted pendulum are shown in Figure \ref{fig:strategy_corr}. A paired t-test shows that MSRD achieves a statistically significantly higher correlation (M = 0.779, SD = 0.239) for the strategy rewards versus AIRL (M = 0.635, SD = 0.324) trained separately for each strategy, $t(19) = 1.813$, $p = 0.0428$ (one-tailed). A Shapiro-Wilk test showed the residuals were normally distributed ($p = 0.877$). For the hopper domain, MSRD achieved $0.85$ and $0.93$ correlation coefficient for the hop and crawl strategy, compared with AIRL's $0.80$ and $0.82$ respectively. We omit a t-test here due to the limited number of strategies. We could test the discrimination of strategy rewards by evaluating each strategy's reward function on each strategy's trajectory; we expect to observe that the strategy-only reward function of each strategy gives its corresponding trajectory the highest reward. We show in Figure \ref{fig:ip_heatmap} that this expectation holds. Out of $20$ strategy-only rewards, $16$ receive highest rewards in corresponding trajectories. A Binomial test shows we are significantly better than chance ($p < .001$). 

We are unable to examine the reward landscape in the inverted pendulum environment as it is four dimensional. Thus, we choose to fix three dimensions (cart position, cart velocity and pendulum angular velocity) to zero and investigate the reward change within the one remaining dimension (pendulum angle). The relationship between rewards and pendulum angles in task and strategy reward functions are illustrated in Figure \ref{fig:ip_reward_f}, in which the task reward function reaches its peak when the angle of the pendulum is near zero. This precisely recovers the task reward. For strategy-only reward functions, strategy 13 encourages the pendulum to lean left (demonstration behavior shown in bottom-left of Figure \ref{fig:ip_reward_f}), while strategy 4 encourages the policy to tilt the pendulum right (demonstration behavior shown in bottom-right). Therefore, strategy-only rewards learned by MSRD captures specific preferences within demonstrations. Figure \ref{fig:ip_reward_f} also shows the magnitude of the task reward is larger than the strategy reward, which affirms our expectation that an emphasis is being put towards accomplishing the task.

\begin{figure}[b]
  \centering
  \includegraphics[width=\linewidth]{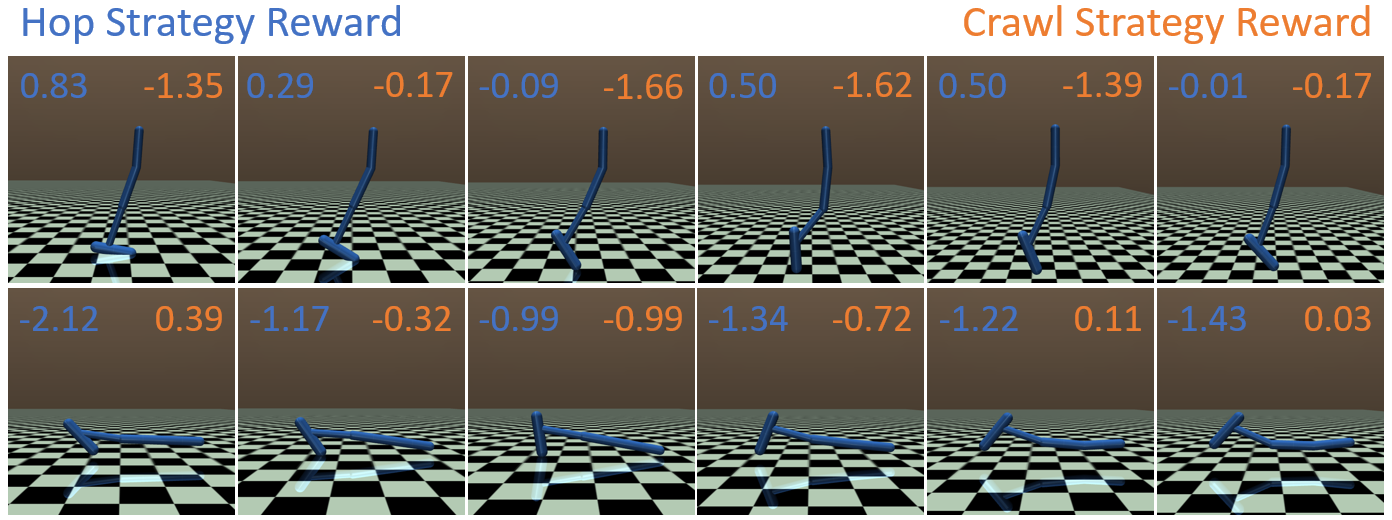}
  \caption{Hop and crawl strategy reward on Hop (top) and Crawl (bottom) trajectories. Blue and orange numbers are learned Hop and Crawl strategy rewards, respectively.}
  \Description{}
  \label{fig:hopper_strategy_reward}
\end{figure}

In the hopper environment, it is harder to visualize the reward landscape due to a high-dimensional observation space and a lack of interpretability of states. Therefore, instead of visualizing a reward curve, we evaluate the estimated strategy-only reward on trajectories from both strategies to provide evidence for \textbf{H2}. Figure \ref{fig:hopper_strategy_reward} shows that when given a hopping trajectory, the hop strategy-only reward function gives higher reward for that behavior than crawl strategy-only reward function. Similarly, in the crawl trajectory case (Figure \ref{fig:hopper_strategy_reward}), the crawling strategy-only reward gives a higher value than the hop strategy-only reward. Therefore, the strategy-only reward function recovered by MSRD gives a higher reward to the corresponding behavior than the other strategy-only reward function,  thus providing encouragement to the policy towards the intended behavior (\textbf{H2}).

These results across our simulated environments show our algorithms' success in both task reward recovery and strategy reward decomposition. This capability is a novel contribution to the field of LfD in that we are able to tease out strategies from the underlying task and effectively learn policies that can reproduce both the strategy and a well-performed policy for the underlying task. 

\begin{figure}[b]
  \centering
  \includegraphics[width=\linewidth, height = 6cm]{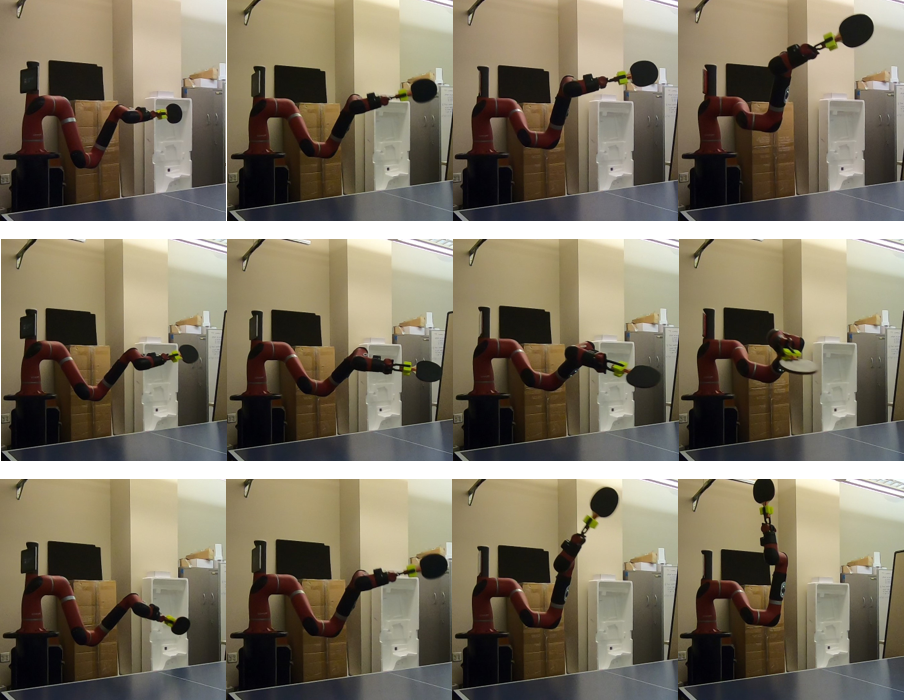}
  \caption{Different strategies learned in robot table tennis task. Top to bottom: Push, Slice, and Topspin. }
  \Description{}
  \label{fig:pingpong}
\end{figure}
\subsection{Physical Environment}
\begin{figure}[b]
  \centering
  \includegraphics[width=0.8\linewidth, height = 4.5cm]{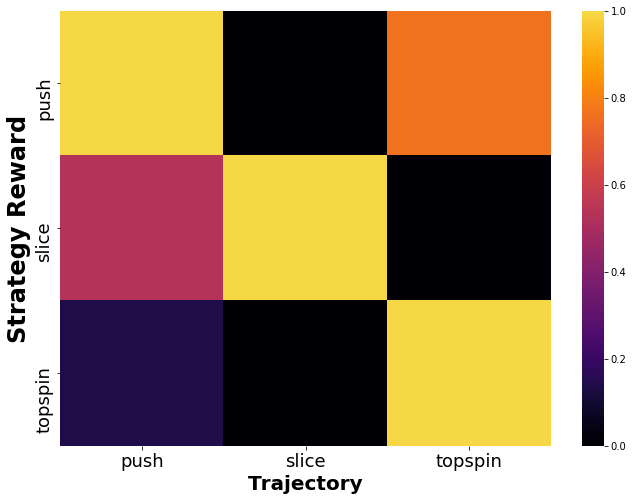}
  \caption{Evaluations of Strategy Rewards on Strategy Trajectories for Table Tennis. Each row is normalized to $[0,1]$. }
  \Description{}
  \label{fig:matrix_pingpong}
\end{figure}

We utilize four deep neural networks (DNNs) consisting of three fully connected layers (32 nodes on each hidden-layer) to represent task, push, slice and topspin rewards. The ball states alongside robot arm joints serve as our inputs. The label of different types of demonstration (forehand-push, backhand-slice, etc.) is available to our algorithm. Figure \ref{fig:pingpong} shows four frames of the learned trajectories for our defined strategies. The change in angle and the upward/downward motion of the paddle throughout the policy trajectory are key factors in the display of different strategies (as these induce spin). Push is associated with a small change in angle as it is not attempting to add any spin onto the ball. Slice places a backspin on the ball, and thus the angle of the paddle will quickly tilt up as shown in Figure \ref{fig:pingpong}. Conversely, topspin places a topspin on the ball; to do so, the associated trajectory has a quick upward motion. Figure \ref{fig:matrix_pingpong} provides quantitative evidence that the strategy-only reward should be maximal given a demonstration utilizing the corresponding strategy. After just 30 minutes of training on each strategy, the robot was able to learn to strike $83\%$ of the fed balls. The robot learned to perform all strategies, and the robot's best strategy, topspin, resulted in $90\%$ of balls returned successfully.

To further verify that our task reward was learning the correct behavior, the task reward function was used to evaluate each of the demonstrated trajectories (three trajectories for each of the three strategies). In the case of the original demonstrations (i.e., where the ball was struck and landed in bounds), the average and standard deviation of the task reward across demonstrations and strategies was $1.476 \pm 0.051$. We then virtually manipulated the trajectories to indicate that the ball was unsuccessfully returned, which achieved an average and standard deviation for the task reward of only $1.284 \pm 0.042$. A Friedman Test shows this result is statistically significant ($\chi^2(1,9)=9, p < 0.05$), providing support that our task reward is learning to identify successful returns, as unsuccessful trajectories should be associated with lower reward when compared to successful. A video of the robot returning an incoming ball can be found in our supplementary materials. In this real-world robot control task, we see that MSRD can successfully recover all three strategies' behaviors using human data.

\section{Future Work} Our approach to teasing out demonstrator-specific rewards from a common, task reward might be able to tease out sub-optimality. Researchers have discussed models for human sub-optimality, e.g., $\epsilon$-greedy approaches~\cite{brown2019ranking,gombolay2016apprenticeship}. However, there is an argument to be made for sub-optimality being reflected via satisficing heuristics~\cite{klein1993decision,simon1972theories} or a person-specific augmentation of a task reward~\cite{vinyals2019grandmaster}, which is analogous to MSRD's model. In future work, we aim to explore models of human decision-making towards being able to recover a demonstrator's latent objective from sub-optimal demonstration. We also plan to utilize different inference methods to allow for unlabelled demonstrations (i.e., to infer the strategy label).

\section{Conclusion}
We explored a new problem of apprenticeship via IRL when demonstrators provide heterogeneous, strategy-based demonstrations seeking to maximize a latent task reward. By explicitly modeling the task and strategy reward separately, and jointly learning both, our algorithm is able to recover more robust task rewards, discover unique strategy rewards, and imitate strategy-specific behaviors. 
\begin{acks}
We want to thank Dr.~Sonia Chernova and the RAIL lab for letting us borrow their Sawyer robot and for Dr.~Chernova's valuable feedback in revising our manuscript. This work was sponsored by ONR under grant N00014-18-S-B001, MIT Lincoln Laboratory grant 7000437192, and GaTech institute funding.
\end{acks}

\bibliographystyle{ACM-Reference-Format}
\bibliography{sample-sigconf}


\end{document}